\documentclass{llncs}

\usepackage{graphicx}
\usepackage{amsmath}
\usepackage{amssymb}
\usepackage{multirow}
\usepackage{layouts}
\usepackage{cite}
\usepackage{breqn}
\usepackage{subfig}
\usepackage{subfloat}

\begin{document}
\title{Position-Aware Convolutional Networks for Traffic Prediction}

\author{Shiheng Ma\inst{1} \and
Jingcai Guo\inst{2} \and
Song Guo\inst{2} \and Minyi Guo\inst{1}}

\authorrunning{S. Ma et al.}

\institute{
Shanghai Jiao Tong University, Shanghai 200240, China \\
\email{ma-shh@sjtu.edu.cn, guo-my@cs.sjtu.edu.cn} 
\and
The Hong Kong Polytechnic University, Hong Kong, China \\
\email{cscjguo@comp.polyu.edu.hk, song.guo@polyu.edu.hk}
}

\maketitle



\begin{abstract}
Forecasting the future traffic flow distribution in an area is an importance issue for traffic management in an intelligent transportation system. 
The key challenge of traffic prediction is to capture spatial and temporal relations between future traffic flows and historical traffic due to highly dynamical patterns of human activities. 
Most existing methods explore such relations by fusing spatial and temporal features extracted from multi-source data.
However, they neglect position information which helps distinguish patterns on different positions. 
In this paper, we propose a position-aware neural network that integrates data features and position information. 
Our approach employs the inception backbone network to capture rich features of traffic distribution on the whole area. 
The novelty lies in that under the backbone network, we apply position embedding technique used in neural language processing to represent position information as embedding vectors which are learned during the training. 
With these embedding vectors, we design position-aware convolution which allows different kernels to process features of different positions. 
Extensive experiments on two real-world datasets show that our approach outperforms previous methods even with fewer data sources. 
\end{abstract}

\section{Introduction} \label{sec: introduction}

Traffic prediction, forecasting the future traffic distribution in a city by its historical traffic records, plays an important role in multiple domains. 
It is a crucial help to build an intelligent transportation system for urban planning and development \cite{DBLP:journals/tist/ZhengCWY14a}. 
Knowing the future traffic distribution, we can make alternative traffic plans for traffic congestions before they occur. 

Traffic prediction has been studied for years. 
In the early days of the study, time series models, such as autoregressive integrated moving average (ARIMA), have been widely used \cite{shekhar2007adaptive,li2012prediction,moreira2013predicting,lippi2013short}. 
These models focus on extracting temporal patterns of changing traffic in one area. 
Later on, the spatial correlation of different areas has been investigated by integrating traffic data in multiple areas to make predictions.
Moreover, additional data, e.g., weather, date, is integrated into prediction models \cite{pan2012utilizing,wu2016interpreting,rong2017taxi}. 
Recently, the introduction of neural networks sharply improves the prediction accuracy. 
Neural models can effectively capture temporal-spatial features from traffic data and fuse additional features in other data sources by using various techniques, e.g., convolutional neural networks (CNN), long short term memory (LSTM), attention mechanisms, etc.
Our work shows there is still room for improvement by exploiting position information and feature diversity. 

\begin{figure}[b]
    \centering
    \includegraphics{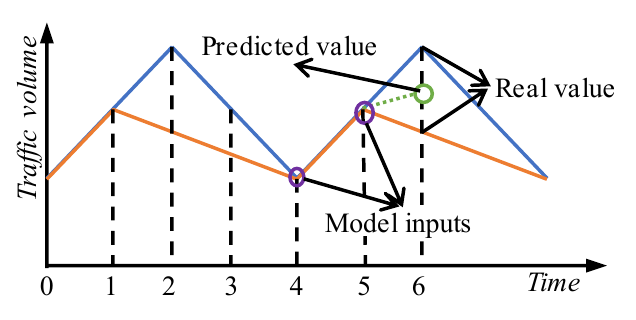}
    \caption{
        Prediction failure caused by the same model for different positions. The blue and red orange lines represent the traffic volumes in two positions, respectively. 
    }
    \label{fig: model diff}
\end{figure}

\textit{Position information}. 
Most previous works focus on extracting features of traffic data itself and neglect the position information. 
Specifically, for traffic data in different positions, their position information is unknown to the model and does not influence the parameters of the model, i.e., different positions share the same model. 
This is because these methods apply image processing techniques, e.g., CNN with kernels of shared parameters, which aim to find common features of overall figures. 
Different from image processing, the pattern difference of positions cannot be neglected in traffic prediction and it is difficult to capture position features from data. 
Fig.~\ref{fig: model diff} shows a simple case that uses one model for two positions. 
Suppose that the model predicts future value by recent two data point, if the position information is unknown to the model, the model will give the same predicted value for the two positions. 
Thus, we can see that position information is an intrinsic property which does not depend on traffic data. 
Some recent works use additional information and can mitigate the problem but 
they cannot essentially overcome the difficulty because they do not explicitly take position information into account.

\textit{Feature diversity}. 
Recent neural network models have considered multi-source data and various spatial-temporal dependencies between future traffic distributions and historical traffic distributions. 
However, they neglect feature diversity, i.e., using only one kind of features. 
Specifically, these models make predictions based on features extracted by the same network. 
Neither spatial nor temporal patterns can be clearly described by one kind of features since these patterns may contain multiple abstraction levels, different frequencies, etc.
This is similar to signal representation: a set of bases of the same frequency cannot represent a signal composed of a number of different frequencies. 

To address the two issues above, we propose a position-aware network model (PAN). 
The framework of PAN borrows the idea of video frame prediction \cite{DBLP:journals/corr/MathieuCL15}. 
Specifically, PAN takes historical traffic distributions as a series of images. 
Based on these images, PAN generates a new image as the predicted traffic distribution. 
This framework allows extracting spatial-temporal features of all positions at the same time. 
It benefits to learn comprehensive features. 

Under the framework, we design a position embedding mechanism based on representation learning to capture the position information. 
Our position embedding is inspired by word position embedding in neural language processing (NLP) which distinguishes different semantics of one word in different positions of a sentence \cite{DBLP:journals/corr/abs-1810-04805}. 
All positions are embedded into a vector space. 
Embedding vectors indicate features of each position and they are learned with the prediction model. 
Different from the usage in NLP, we use position embedding not only in the input layer but also in convolutional layers to modify the representation of input features. 
Due to the utilization of position information, PAN can provide different prediction patterns for different positions with less information. 

We employ position-aware inception blocks to model spatial-temporal dependencies for extracting diverse features. 
The network structure is based on the inception network \cite{DBLP:conf/aaai/SzegedyIVA17}. 
In each position-aware inception block, we design multiple convolutional layers with different kernel sizes and depths. 
Different convolutional layers can extract different kinds of dependencies between input features and output features
Meanwhile, position embedding is added into convolutional layers. 
These position-aware features in different kinds make PAN more expressive. 

Our contributions are summarized as follows:
\begin{itemize}
    \item 
    We propose a new position-aware network for capturing the position information and promoting the feature diversity, which are two particular properties in traffic prediction but are neglected by most previous studies. 
    \item 
    We design a position embedding mechanism to learn position features simultaneously with prediction model training. 
    Embedding vectors are fused with multiple layers for building different patterns for different positions. 
    \item 
    We employ position-aware inceptions blocks which parallelly use different convolutional sub-network with position embedding to promote feature diversity. 
    \item 
    We evaluate our model on several real-world traffic datasets. 
    The results show that our model requires fewer data and outperforms other state-of-the-art methods.
\end{itemize}

The rest of this paper is organized as follows. 
We first summarize recent studies of traffic prediction in Section 2. 
Then, we formulate the problem of traffic prediction in Section 3. 
Detailed design of the PAN model is presented in Section 4. 
We evaluate the performance of PAN on two real-world datasets in Section 5. 
Finally, we draw the conclusion in Section 6. 

\section{Related Work} \label{sec: related work}

In most recent years, the traffic prediction has gained increasing attention in machine learning and data mining areas. Numerous studies focusing on how to properly utilize the traffic-related data resource and accurately forecast the future traffic flow distribution have been proposed and obtained successive state-of-the-art performances.

Early works \cite{shekhar2007adaptive,li2012prediction,moreira2013predicting,lippi2013short,pan2012utilizing,wu2016interpreting,rong2017taxi,ide2011trajectory,zheng2013time,deng2016latent,tong2017simpler} which utilized time series methods based on statistical learning and classic machine learning have been widely studied. Shashank \textit{et al.}\cite{shekhar2007adaptive} proposed the autoregressive integrated moving average (ARIMA) for traffic prediction. Li \textit{et al.} \cite{li2012prediction} improved the ARIMA to forecast the spatial-temporal variation of passengers in a hotspot. Moreira-Matias \textit{et al.} \cite{moreira2013predicting} aggregated the streaming information and ensembled three time-series forecasting techniques to originate a prediction. Lippi \textit{et al.} \cite{lippi2013short} proposed to use the support vector regression model combining with a seasonal kernel to measure similarity between time-series examples. Some studies \cite{pan2012utilizing,wu2016interpreting,rong2017taxi} extended the prediction to further use some external data resource, such as venue types, weather conditions, event information, etc. Moreover, some methods also embedded the spatial information into the models and obtained some promising results \cite{ide2011trajectory,zheng2013time,deng2016latent,tong2017simpler}. However, these methods require data to satisfy some assumptions or need careful feature engineering. Thus, they usually cannot model too complex data and perform poorly in practice.

Recently, with the rapid development of deep learning \cite{lecun2015deep}, traditional time series methods are inferior to deep learning based methods on multi-level aspects. 
In some studies, the traffic distributions of entire city are treated as images. 
For example, CNN can be directly applied on images of traffic speed for speed prediction \cite{DBLP:journals/sensors/MaDHMWW17}. 
To capture more complex features and increase the depth of neural networks, residual network are proposed to use on traffic flow prediction \cite{DBLP:conf/gis/ZhangZQLY16,DBLP:conf/aaai/ZhangZQ17}. 
Although residual networks perform well in image processing, applying them on traffic prediction need consider the characteristics of the problem. 
Some other works use traffic data of neighbor areas to predict the future traffic of the centric area. 
Most of these works employ both convolutional neural networks (CNN) and long short-term memory networks (LSTM) for capturing spatial and temporal dependencies, respectively. 
For example, Yao \textit{et al.} \cite{DBLP:conf/aaai/Yao0KTJLGYL18} apply an LSTM to integrate the outputs of CNNs. 
Zhou \textit{et al.} \cite{DBLP:conf/wsdm/ZhouSZH18} employ convolutional LSTM which use convolution operation in LSTM units for prediction passenger demands. 
Yao \textit{et al.} \cite{yao2019revisiting} consider the dynamic similarity between locations and propose an attention mechanism for LSTM-connected CNNs. 
Using neighbor traffic data reduces the interference of low-correlation data on prediction results. 
Meanwhile, we note that addition information, e.g., date, weather, traffic flow direction, are widely used in most recent deep learning based model \cite{DBLP:conf/aaai/ZhangZQ17, DBLP:conf/aaai/Yao0KTJLGYL18, yao2019revisiting}. 
Some studies focus traffic flow on roads or traffic on separated nodes which can be seen as graphs. 
In this scenario, graph convolutional network (GNN) are used. 
For example, Wang \textit{et al.} \cite{DBLP:conf/icnp/WangZYLP17} apply a GNN to model spatial features based on in-cell and inter-cell data traffic. 
Guo \textit{et al.} \cite{guo2019attention} add an attention mechanism on GNN to capture dynamic spatial-temporal correlations for road traffic flow prediction. 
This is a different kind of traffic prediction task. 
Thus, we do not discuss GNN-based models in this paper.

\section{Problem Formulation} \label{sec: formulation}

Lots of vehicles record and report their states while moving, e.g., trip starting or ending, moving or stopping, etc.. 
Suppose there are $K$ kinds of states containing geographic coordinates. 
To depict the overall traffic situation, we partition the whole city into $I \times J$ small square cells and divide time into $T$ timeslots. 
Then, we define the traffic distribution as $d^{(t)} \in \mathbb{R} ^{I \times J \times K}$ where $d^{(t)}_{ijk}$ is the number of state $k$ in cell $(i,j)$ at timeslot $t$. 
Accordingly, a temporal sequence $\{ {d}^{(t)}, {d}^{(t-1)}, \cdots, {d}^{(t-L+1)} \}$ with $L$ number of past time slots forms the spatial-temporal demand distribution. 
With this distribution, the traffic prediction problem can be formulated as finding a model $\mathcal{P}$ such that it can minimize the prediction error with an error metric $\mathcal{E}$, i.e., 
$$
    \min_{\mathcal{P}}\; 
    \mathcal{E}
    \left(
        \mathcal{P} \left(
            {d}^{(t)}, {d}^{(t-1)}, \cdots, {d}^{(t-L+1)} 
        \right), {d}^{(t+1)}
    \right),
$$
where the input of $\mathcal{P}$ is $L$-length temporal sequence of recent traffic distribution $\{ {d}^{(t)}, {d}^{(t-1)}, \cdots, {d}^{(t-L+1)} \}$. 
Moreover, the prediction model $P$ predicts the whole future distribution simultaneously. 

\section{Position-Aware Network} \label{sec: model}

\begin{figure}[tb]
    \centering
    \includegraphics{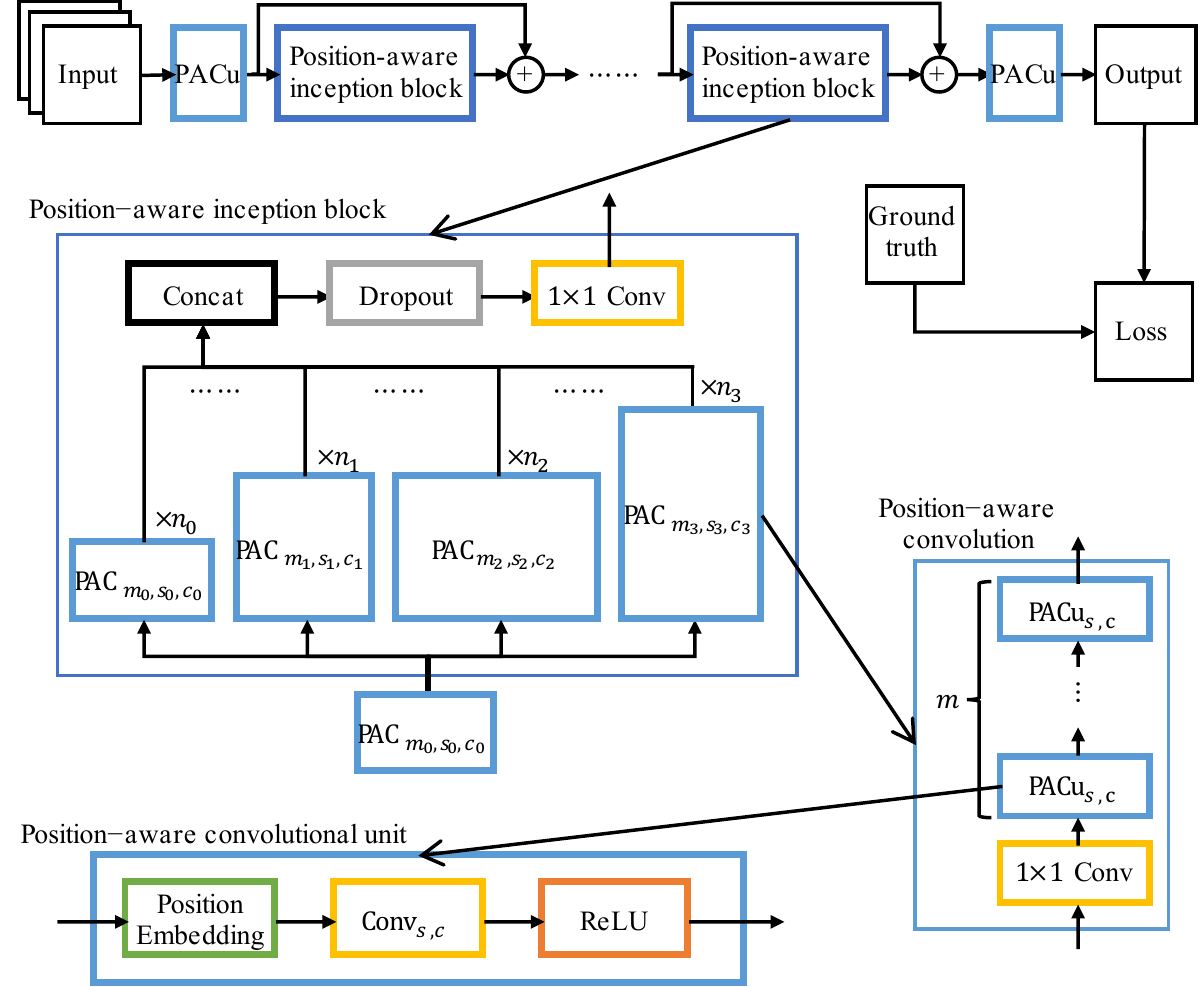}
    \caption{
        The architecture of PAN.  
    }
    \label{fig: arch}
\end{figure}

In this section, we describe the design of our Position-Aware Network (PAN) prediction model $\mathcal{P}$. 
Fig.~\ref{fig: arch} shows the architecture of PAN which is based on Inception-ResNet \cite{DBLP:conf/aaai/SzegedyIVA17} backbone network. 
The input integrates historical traffic distribution and the position embedding features. 
The position-aware spatial-temporal inception (PASTI) blocks are combined with residual connection. 
In each PASTI, multiple position-aware convolutional modules (PAC) in different kinds are parallelly integrated to extract different kinds of features. 
In order to adapt to the traffic prediction problem, we add position embedding layers (PE) followed by convolutional layers in PAC. 
Moreover, we use dropout instead of batch normalization to avoid overfitting since normalization keeps relative values of features and destroy absolute values which are important to prediction tasks. 

\subsection{Input Constructing}

Based on common sense, traffic series have three obvious correlations: recent, daily, and weekly. 
The correlated traffic distributions are most helpful to predict the future distribution. 
Recent correlation indicates the future traffic depends on recent traffic, i.e., future traffic distribution is similar to distributions of recent hours. 
We define the sequence of recent traffic distributions as 
\begin{align*}
    D_{recent} = \left\{ d^{(t)}, d^{(t-1)}, \cdots, d^{(t-n_{r}+1)} \right\} ,
\end{align*}
where hyper-parameter $n_{r}$ is the number of recent timeslots. The length of $D_{recent}$ is $n_{r}$.

Daily correlation indicates the future traffic is related to the traffic at the same time in past days, i.e., today's traffic distribution is similar to yesterday's. 
We define the sequence of traffic distributions in recent days as
\begin{align*}
    D_{daily} = & \left\{ d^{(t+n_{r}-L_{d})}, \cdots, d^{(t-n_{r}+1-L_{d})}, \cdots, \right. \\
    & \left. d^{(t+n_{r} - n_{d} L_{d} )}, \cdots, d^{(t-n_{r}+1 - n_{d} L_{d})} \right\} ,
\end{align*}
where hyper-parameter $n_{d}$ is the number of recent days and constant $L_{d}$ is the number of timeslots in one day. 
Since the periods of recent days which have similar traffic distributions may deviate, we extend $\{ d^{(t+1-L_{d}))}, \cdots, d^{(t-n_{r}+1-L_{d})} \}$ by adding $ \{ d^{(t+n_{r}-L_{d})}, \cdots, d^{(t+2-L_{d}))}  \}$. 
The length of $D_{daily}$ is $2 n_{r} n_{d}$. 

Weekly correlation indicates the future traffic is related to the traffic at the same time in past weeks, i.e., today's traffic distribution is similar to the same day of past weeks. 
We define the sequence of traffic distributions in recent weeks as
\begin{align*}
    D_{weekly} = & \left\{ d^{(t+n_{r}-L_{w})}, \cdots, d^{(t-n_{r}+1-L_{w})}, \cdots, \right. \\
    & \left. d^{(t+n_{r} - n_{w} L_{w} )}, \cdots, d^{(t-n_{r}+1 - n_{w} L_{w})} \right\} ,
\end{align*}
where hyper-parameter $n_{w}$ is the number of recent weeks and constant $L_{w}$ is the number of timeslots in one week. 
The length of $D_{weekly}$ is $2 n_{r} n_{w}$. 

We concatenate all distributions in the three sequences along the last dimension as the input of our model at timeslot $t$:
\begin{align*}
    D_{input}^{(t)} = \operatorname{Concat} \left( D_{recent} \cup D_{daily} \cup D_{weekly} \right) \in \mathbb{R}^{I \times J \times (n_{r} + 2 n_{r} n_{d} + 2 n_{r} n_{w}) K} .
\end{align*}

\subsection{Position Embedding}

Inspired by word position embedding in neural language processing, we employ representation learning to generate feature vectors of positions. 
Positions are embedded into a vector space. 
Each embedding vector represents the information of its corresponding position and it is learned with together with the other parts of PAN. 
Since PAN process the entire distribution at the same time, the entire embedding vectors can be denoted as $E \in \mathbb{R}^{I \times J \times f}$ where $f$ is the length of one embedding vector. 
Then, we build position embedding layer $\operatorname{PE}$ which fuses the input features $F \in \mathbb{R}^{I \times J \times c_{F}}$ and position information:
\begin{align*}
    \operatorname{PE} (F) = F + E \in \mathbb{R}^{I \times J \times c_{F}}. 
\end{align*}
where embedding vectors $E$ has the same shape of $F$ and $c_{F}$ is the number of channel of $F$. 
Here, we follow the fusing approach in BERT \cite{DBLP:journals/corr/abs-1810-04805}. 
We have tried other fusing approaches such as multiplication or concatenation. 
However, the sum fusing achieves the best results. 

We use $\operatorname{PE}$s not only in the in the input, but also in multiple parts of PAN. 
Since features in different parts have different meanings, these $\operatorname{PE}$s should not share a common $E$. 
Thus, we give each $\operatorname{PE}$ independent parameters $E$.

\subsection{Position-Aware Convolutional Modules}

In order to capture spatial and temporal features for different positions, we build position-aware convolutional modules (PAC). 
PAC is a stack of PAC units. 
A PAC unit (PACu) is composed of three components: position embedding, 2D convolution and ReLU activation \cite{DBLP:conf/icml/NairH10}. 
Given input features $F$, PACu transform $F$ by the following formulation:
\begin{align*}
    \operatorname{PACu}_{s, c} ( F ) = \operatorname{ReLU}\left(\operatorname{Conv}_{s, c}\left(\operatorname{PE}\left(F\right)\right)\right) \in \mathbb{R}^{I \times J \times c},
\end{align*}
where PACu has two hyper-parameters, $s$ and $c$, respectively indicating the kernel size and the number of filters (output channels) of the convolutional layer in PACu. 
$\operatorname{Conv}_{s, c} (F) = W * F + b$ is a convolution with kernels $W \in \mathbb{R}^{c \times s \times s}$ and bias $b \in \mathbb{R}^{c}$. 
Meanwhile, $W$ and $b$ are learned parameters in PACu. 
$\operatorname{ReLU}(x) = \max \{x,0\}$ is an element-wise function and does not change the shape of input features. 

Then, we define PAC as
\begin{align*}
    \operatorname{PAC}_{m,s,c}(F) = \operatorname{Conv}_{1, c} (
        \overbrace{
            \operatorname{PACu}_{s, c} ( \cdots \operatorname{PACu}_{s, c} 
        }^{\text{$m$ PACu}_{s, c}} 
        (\operatorname{Conv}_{1,c} (F) ) )
    ) \in \mathbb{R}^{I \times J \times c} ,
\end{align*}
where a $\operatorname{PAC}_{m,s,c}$ has three hyper-parameters: $m$ is the number of PACu in PAC, $s$ and $c$ are defined in PACu. 
The depth and width of $\operatorname{PAC}_{m,s,c}$ can be adjusted by setting up $m$ and $c$. 
This makes PAN capture different kinds of features. 

\subsection{Position-Aware Spatial-Temporal Inception Blocks}

To increase the diversity of feature, we compose multiple PACs in a position-aware spatial-temporal inception blocks (PASTI). 
In this paper, we select three kinds of PACs: $\operatorname{PAC}_{1,1,c_{0}}$, $\operatorname{PAC}_{1,3,c_{1}}$, and $\operatorname{PAC}_{2,3,c_{2}}$. 
And the numbers of the three kinds are $n_{0}$, $n_{1}$, and $n_{2}$, respectively. 
Then, we define PASTI as 
\begin{align*}
    \operatorname{PASTI}(F) & = \operatorname{ReLU}(\operatorname{Conv}_{1, c_{F}} ( \operatorname{Dropout} ( \operatorname{Concat}( \\
        &\operatorname{PAC}_{1,1,c_{0}}^{(1)}(F), \cdots, \operatorname{PAC}_{1,1,c_{0}}^{(n_{0})}(F), \\
        &\operatorname{PAC}_{1,3,c_{1}}^{(1)}(F), \cdots, \operatorname{PAC}_{1,3,c_{1}}^{(n_{1})}(F), \\
        &\operatorname{PAC}_{2,3,c_{2}}^{(1)}(F), \cdots, \operatorname{PAC}_{2,3,c_{2}}^{(n_{2})}(F) ) ) ) )
        \in \mathbb{R}^{I \times J \times c_{F}},
\end{align*}
where $c_{F}$ is the number of channels of input features $F$. 
$\operatorname{Concat}$ concatenates all outputs of PACs along the last dimension. 
$\operatorname{Dropout} (X) = M \odot X$ is used during the training to address overfitting problem of deep neural networks \cite{DBLP:journals/jmlr/SrivastavaHKSS14}. 
$\odot$ is element-wise product.
$M$ is a random binary matrix where zero indicate drop a neural unit of $X$.  
The conventional approach to address overfitting in convolutional layers is batch normalization. 
However, we find batch normalization of features decrease the performance of PAN. 

Deep neural networks have advantage of capturing features but it is difficult to train. 
Thus, we apply residual connection \cite{DBLP:conf/cvpr/HeZRS16} in PASTIs as follows
\begin{align*}
    F' = \operatorname{PASTI}(F) + F. 
\end{align*}

\subsection{Loss Function and Training}

To obtain the final prediction results, we employ a $\operatorname{PACu}_{1,K}$ after the last PASTI as shown in Fig.~\ref{fig: arch}. 
The output of PAN is defined as 
\begin{align*}
    \hat{d}^{(t+1)} = \mathcal{P}(D_{input}^{(t)}) \in \mathbb{R}^{I \times J \times K}.
\end{align*}

Usually, mean average percentage error (MAPE) and rooted mean square error (RMSE) are used to measure the accuracy of prediction results. 
MAPE gives higher weights for errors which true values are small while RMSE gives higher weights for larger errors. 
Thus, we combine the two error measurement in our loss:
\begin{align*}
    \mathcal{E} (\hat{d}^{(t+1)}, {d}^{(t+1)}) = \left\| (\hat{d}^{(t+1)} - {d}^{(t+1)})(1-{d}^{(t+1)}) \right\| _1 + \left\| (\hat{d}^{(t+1)} - {d}^{(t+1)}) \right\| _2 ^2,
\end{align*}
where $\|\cdot\|_1$ and $\|\cdot\|_2$ are 1-norm and 2-norm of a matrix, respectively.

\section{Experiments} \label{sec: evaluation}

\subsection{Settings}

\subsubsection{Dataset description}
We evaluate our model with other baselines on two public real-world datasets from New York City: BikeNYC and TaxiNYC. 
For comparison, we select the same datasets used in \cite{yao2019revisiting} and adopt the same settings. 
The two datasets contains trips of renting bikes or taking taxis. 
There are two states of a trip: start and end
Each trip records the time and the coordinate of the trip starting and ending. 
The detail of the two datasets are shown in Table~\ref{tab: datasets}. 
The whole city is partitioned into $10\times20$ regions with the size of around $1km\times1km$. 
Each day is split into 48 timeslots and the length of each timeslot is 30 minutes. 
Both of them have trip data of 60 days. 
In this paper, we use the first 40 days as the training set and the last 20 days as test set.

\begin{table*}[b]
    \caption{Statistics of datasets}
    \label{tab: datasets}
    \centering
    \begin{tabular*}{1.\textwidth}{@{\extracolsep{\fill} } cccccc}
        \hline
        & Starting time & Ending time & \# trips & \# states & Area size \\
        \hline
        BikeNYC & 2016.07.01 & 2016.08.29 & 2,605,648 & 2 &  $10\times20$\\
        TaxiNYC & 2015.01.01 & 2015.03.01 & 22,349,490 & 2 &  $10\times20$\\
        \hline
        \hline
        & \multicolumn{2}{c}{Training set} & \multicolumn{2}{c}{Test set} & Time interval \\
        \hline
        BikeNYC & \multicolumn{2}{c}{2016.07.01 - 2016.08.09} & \multicolumn{2}{c}{2016.08.09 - 2016.08.29} & 30 min \\
        TaxiNYC & \multicolumn{2}{c}{2015.01.01 - 2015.02.10} & \multicolumn{2}{c}{2015.02.11 - 2015.03.01} & 30 min \\
        \hline
    \end{tabular*}
\end{table*}

\subsubsection{Evaluation Metric}
We use rooted mean square error (RMSE) and mean average rercentage error (MAPE) are two most common metrics to compare our model with other baselines. 
For each kind of states, we compute their errors respectively. 
Given prediction results $\hat{d}^{(t)}$ from timeslot $T_1+1$ to timeslot $T_2$ and corresponding ground truth $d^{(t)}$, the two metrics are defined as 
\begin{align*}
    \operatorname{RMSE}_k &= \sqrt{ 
        \frac{1}{(T_2 - T_1)IJ} 
        \sum_{t=T_1+1}^{T_2} 
        \sum_{i=1}^{I} 
        \sum_{j=1}^{J} 
        (\hat{d}_{ij}^{(t)} - d_{ijk}^{(t)})^2 
        } , \\
    \operatorname{MAPE}_k &= \frac{1}{(T_2 - T_1)IJ} \sum_{t=T_1+1}^{T_2} \sum_{i=1}^{I} \sum_{j=1}^{J} \left|\frac{\hat{d}_{ijk}^{(t)} - d_{ijk}^{(t)}}{d_{ijk}^{(t)}}\right| .
\end{align*}
In the two datasets, there are two volumes to predict: the number of trip starting (Start) and the number of trip ending (End). 
Meanwhile, we follow the filtering settings in \cite{yao2019revisiting,DBLP:conf/aaai/Yao0KTJLGYL18}. 
The samples with volume values less than 10 are filtered out since people have little interest of low traffic in the real-world applications. 
Moreover, prediction results on low traffic usually have small RMSE and make MAPE failure (small value as denominator). 

\begin{table*}[t]
\newcommand{\tabincell}[2]{\begin{tabular}{@{}#1@{}}#2\end{tabular}}
    \caption{Evaluation results on TaxiNYC}
    \label{tab: taxi}
    \centering
    \begin{tabular*}{0.85\textwidth}{@{\extracolsep{\fill} } cccccccc}
        \hline
        \multirow{2}{*}{TaxiNYC}&   \multicolumn{2}{c}{Start}  & \multicolumn{2}{c}{End} & \multirow{2}{*}{\tabincell{c}{Additional\\Information}} \\
        \cline{2-5}
                                & RMSE & MAPE         &  RMSE   &    MAPE &  \\
        \hline\hline
        HA   &   43.82   &  23.18\% & 33.83& 21.14\% & No \\
        LR   &   28.51   &   19.94\% & 24.38& 20.07\% & No \\
        ARIMA   &   36.53   &   22.21\% & 27.25& 20.91\% & No \\
        MLP   &   26.67   &   18.43\% & 22.08& 18.31\% & No \\
        XGBoost   &   26.07   &   19.35\% & 21.72 & 18.70\% & No \\
        LinUOTD   &   28.1   &   19.91\% & 24.39 & 20.03\% & No \\
        ConvLSTM   &   28.48   &   20.50\% & 23.67 & 20.70\% & No \\
        DeepSD   &   26.35   &   18.12\% & 21.95 & 18.15\% & No \\
        ST-ResNet   &   26.23   &   21.13\% & 21.63 & 21.09\% & Yes \\
        DMVST-Net   &   25.74   &   17.38\% & 20.51 & 17.14\% & Yes \\
        STDN        &   24.10    &   16.30\%   &   19.05   &16.25\% & Yes \\
        \hline
        \textbf{PAN}    &   \textbf{21.46}   &   \textbf{14.23\%}   &   \textbf{10.75}   &    \textbf{15.68\%} & No  \\
        \hline
    \end{tabular*}
\end{table*}
\subsubsection{Baselines}
We compare PAN with 11 baselines including both traditional approaches and recent deep neural network models. 
Baselines are historical average (HA), Aautoregressive integrated moving average (ARIMA), ridge regression (ridge), LinUOTD \cite{tong2017simpler}, XGBoost \cite{DBLP:conf/kdd/ChenG16}, multiLayer perceptron (MLP), convolutional LSTM (ConvLSTM) \cite{DBLP:conf/nips/ShiCWYWW15}, DeepSD \cite{DBLP:conf/icde/WangCLY17}, deep spatio-temporal residual networks (ST-ResNet) \cite{DBLP:conf/aaai/ZhangZQ17}, Deep Multi-View Spatial-Temporal Network (DMVST-Net) \cite{DBLP:conf/aaai/Yao0KTJLGYL18}, and spatial-temporal dynamic network (STDN) \cite{yao2019revisiting}. 

\subsubsection{Hyperparameter Settings}
To construct input series, we set length of recent, daily, and weekly series as $n_r = 5$, $n_d = 2$, and $n_w=1$. 
Before training, we normalize traffic data to $[0,1]$ by Min-Max normalization. 
The prediction results will be detransformed for evaluation. 
In PAN, the number of PASTIs is set as 10. 
In each PASTI, the numbers of the three PACs are respectively set as $n_0=1$, $n_1=4$, and $n_2=4$. 
And their numbers of filters are set as $c_0=256$,  $c_1=16$, and $c_2=16$. 
The number of filters in other convolutional layers is set as 256. 
The dropout rate is set as 0.5. 
In training, the batch size is set to 32. 
Learning rate is set as 0.00001. 
Moreover, we use the same hyperparameter settings for training on the two datasets. 

\subsection{Results}

\subsubsection{Prediction Performance}
We compare PAN with all baselines with metrics of RMSE and MAPE. 
As we use the same datasets and evaluation strategy, we directly reuse the results of several baselines from the literature \cite{yao2019revisiting}.

Evaluation results on TaxiNYC are shown in Table~\ref{tab: taxi}. 
PAN significantly outperforms all baselines on TaxiNYC. 
Especially, PAN dramatically improves RMSE of End prediction. 
Meanwhile, the MAPE improvement of End prediction is small. 
This indicates that it is difficult to balance MAPE and RMSE. 
Considering MAPE gives high weights for small ground-truth values and RMSE gives small weights for small errors, we infer that PAN works better at prediction high traffic volumes than low traffic volumes. 
From the results of all baselines, we can find the large performance gap between traditional time series models and recent neural network models. 
However, the performance difference of neural network models is small. 
Comparing the reults of all models, we can find the it is more difficult to improve MAPE than RMSE. 
Moreover, recent nueral network models introduce additional information such as date, weather, and volumes of traffic flow from one area to another. 
Our model only use traffic distribution information.

Evaluation results on TaxiNYC are shown in Table~\ref{tab: bike}. 
PAN significantly outperforms all baselines on TaxiNYC except MAPE of End prediction. 
This is the price of low RMSE of End prediction. 
The results of Start prediction are contrary: small improvement on RMSE and large improvement on MAPE. 
Comparing all results on BikeNYC and TaxiNYC, RMSEs on TaxiNYC are higher than BikeNYC. 
This means the values of Start and End traffic in TaxiNYC are much larger. 
Thus, we can infer that PAN performs better for high traffic than low traffic. 
All models have the same phenomenon since all MAPEs on TaxiNYC are better than BikeNYC.

\begin{table*}[t]
\newcommand{\tabincell}[2]{\begin{tabular}{@{}#1@{}}#2\end{tabular}}
    \caption{Evaluation results on BikeNYC}
    \label{tab: bike}
    \centering
    \begin{tabular*}{0.85\textwidth}{@{\extracolsep{\fill} } cccccccc}
        \hline
        \multirow{2}{*}{BikeNYC}&   \multicolumn{2}{c}{Start}  & \multicolumn{2}{c}{End} & \multirow{2}{*}{\tabincell{c}{Additional\\Information}}  \\
        \cline{2-5}
                                & RMSE & MAPE         &  RMSE   &    MAPE &  \\
        \hline\hline
        HA   &   12.49 &   27.82\% & 11.93 & 27.06\% & No \\
        ARIMA   &   11.53 &   26.35\% & 11.25 & 25.79\% & No \\
        LR   &   10.92 &   25.29\% & 10.33 & 24.58\% & No \\
        MLP   &   9.83 &   23.12\% & 9.12 & 22.40\% & No \\
        XGBoost   &   9.57 &   23.52\% & 8.94 & 22.54\% & No \\
        LinUOTD   &   11.04 &   25.22\% & 10.44 & 24.44\% & No \\
        ConvLSTM   &   10.40  &   25.10\% & 9.22 & 23.20\% & No \\
        DeepSD   &   9.69  &   23.62\% & 9.08 & 22.36\% & No \\
        ST-ResNet   &   9.80   &   25.06\% & 8.85 & 22.98\% & Yes \\
        DMVST-Net   &   9.14   &   22.20\%   &   8.50   & 21.56\% & Yes \\
        STDN        &   8.85    &   21.84\%   &   8.15   & \textbf{20.87\%} & Yes \\
        \hline
        PAN    &   \textbf{8.36}   &   \textbf{19.87\%}   &   \textbf{6.86}   &   23.68\% & No  \\
        \hline
    \end{tabular*}
\end{table*}


\subsubsection{Model Analysis} 
\begin{figure}[bt]
    \centering
    \subfloat[NRMSE. ]{
        \includegraphics{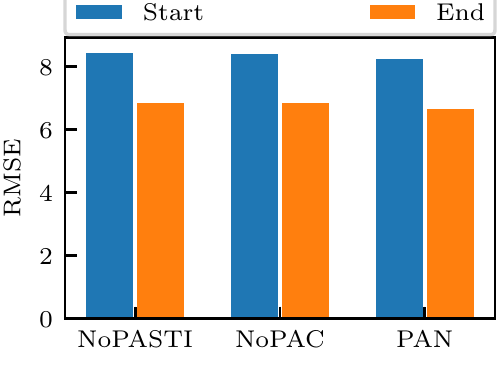}\label{subfig:a}
    }
    \quad
    \subfloat[MAPE. ]{
        \includegraphics{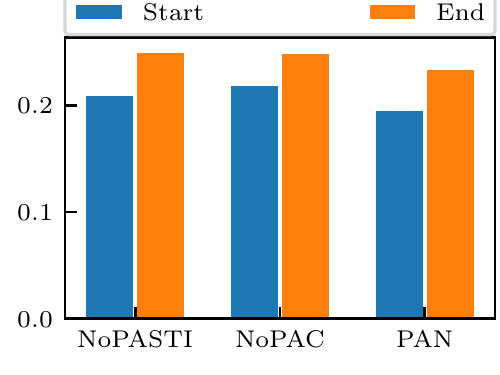}\label{subfig:b}
    }
    \caption{NRMSE/MAPE comparison of various versions of PAN on BikeNYC. }\label{fig: version bike}
\end{figure}

In this section, we study the influence of the position-aware mechanism and diverse features. 
We design two simplified versions of PAN: 
\begin{itemize}
    \item PAN\_NoPAC: Use traditional convolutional layers instead of PAC. 
    \item PAN\_OnePAC: Use only one kind of PAC in each PASTI. 
\end{itemize}
We evaluate them with PAN. 
The results are shown in Fig.~\ref{fig: version bike}
We can see the performance decrease on both datasets. 
This indicates the effectiveness of position information and diverse features.


\section{Conclusion} \label{sec: conclusion}

In this paper, we propose an position-aware network model for the traffic prediction. 
We employ the position embedding mechanism to extract intrinsic information of positions. 
With position embedding, we design the position-aware spatial-temporal inception blocks to capture different kinds of features. 
Position information determines the model can perform differently for different positions. 
Various kinds of features extend the expressiveness of the model. 
Thus, our model outperforms baselines in experiments of two public real-world datasets. 
Extensive experimental results on two public real- world datasets show that our model achieves markable improvements without additional information against the baselines.

\bibliographystyle{splncs04} 
\bibliography{main,related_work}
\end{document}